\documentclass[preprint,12pt]{elsarticle}
\usepackage[utf8]{inputenc}
\usepackage{graphicx}
\usepackage{geometry}
\usepackage{amsmath}
\usepackage{amssymb}

\journal{Engineering Applications of Artificial Intelligence}

\newif\ifanonymous
\anonymousfalse

\begin{document}

\IfFileExists{generated/inline_numbers.tex}{
\newcommand{\FOneTenPct}{0.615}
\newcommand{\FOneFull}{0.771}
\newcommand{\FOneGapFullMinusTen}{0.156}
\newcommand{\FOneSynthThreeX}{0.739}
}{}

\begin{frontmatter}

\title{Synthetic Defect Image Generation for Power Line Insulator Inspection Using Multimodal Large Language Models}

\ifanonymous
\else
\author{Xuesong Wang}
\ead{xswang@wayne.edu}
\author{Caisheng Wang\corref{cor1}}
\ead{cwang@wayne.edu}

\cortext[cor1]{Corresponding author}

\affiliation{organization={Department of Electrical and Computer Engineering, Wayne State University},
            addressline={42 W. Warren Ave.},
            city={Detroit},
            postcode={48201},
            state={MI},
            country={USA}}
\fi

\begin{abstract}
Utility companies increasingly rely on drone imagery for post-event and routine inspection, but training accurate defect-type classifiers remains difficult because defect examples are rare and inspection datasets are often limited or proprietary. We address this data-scarcity setting by using an off-the-shelf multimodal large language model (MLLM) as a training-free image generator to synthesize defect images from visual references and text prompts. Our pipeline increases diversity via dual-reference conditioning, improves label fidelity with lightweight human verification and prompt refinement, and filters the resulting synthetic pool using an embedding-based selection rule based on distances to class centroids computed from the real training split. We evaluate on ceramic insulator defect-type classification (shell vs.\ glaze) using a public dataset with a realistic low training-data regime (104 real training images; 152 validation; 308 test). Augmenting the 10\% real training set with embedding-selected synthetic images improves test F1 score (harmonic mean of precision and recall) from \FOneTenPct{} to \FOneSynthThreeX{} (20\% relative), corresponding to an estimated 4--5$\times$ data-efficiency gain, and the gains persist with stronger backbone models and frozen-feature linear-probe baselines. These results suggest a practical, low-barrier path for improving defect recognition when collecting additional real defects is slow or infeasible.
\end{abstract}


\end{frontmatter}

\section{Introduction}

Power transmission and distribution systems are critical infrastructure, and their resilience depends on timely detection and remediation of component defects. In practice, utilities conduct routine and event-driven inspections, with post-storm inspection playing a key role in rapid fault localization and service restoration \cite{Nguyen2018UAVReview,Shen2020ServiceRestoration}. Recent advances in Unmanned Aerial Vehicles (UAVs) and high-resolution imaging have significantly increased inspection coverage and frequency, but they also create a new bottleneck: large volumes of imagery that are difficult to review manually \cite{Nguyen2018UAVReview}.

Among transmission-line components, insulators are safety-critical because they provide electrical insulation and ensure mechanical support. Ceramic (porcelain) disc insulators are widely deployed and can degrade over time due to weathering and mechanical stress, exhibiting defects such as cracks, chips, and surface damage. Automated insulator defect detection and classification has therefore received increasing attention in the power-systems and computer-vision communities \cite{Liu2023InsulatorSurvey,Liao2019StudyOP}.

Deep learning methods are a natural fit for this setting, but their effectiveness depends strongly on access to labeled defect data. In real deployments, defect examples are intrinsically rare, and inspection datasets are often limited in size and sometimes not publicly available \cite{Nguyen2018UAVReview,Liu2023InsulatorSurvey}. Consequently, a utility may have only a small number of labeled defect samples for a given defect type, making it difficult to train robust defect classifiers and limiting the speed at which new models can be deployed.

A common response to data scarcity is to use data augmentation and transfer learning. Classical image augmentation can improve robustness to nuisance variations (e.g., viewpoint, illumination) by applying label-preserving transformations to existing images \cite{Shorten2019DataAugSurvey,Cubuk2020RandAugment}. However, because it does not add new defect appearance modes, it may offer limited benefit when the primary bottleneck is intra-class diversity of defect appearances. Another direction is to use vision-language models for prompt-based, zero-shot or few-shot recognition, which has been enabled by models such as CLIP (Contrastive Language-Image Pre-Training) \cite{Radford2021CLIP}. However, in industrial defect/anomaly settings, the performance of such approaches can be limited by domain shift and the fact that defects often manifest as subtle, localized cues \cite{Zhou2024AnomalyCLIP,Gu2023AnomalyGPT,Zhang2023AeBAD}.

Generative synthesis offers a third path: create new defect examples to augment the training set. Prior work has explored GAN (Generative Adversarial Network)-based defect synthesis pipelines for defect inspection \cite{zhang2021defect,Jain2022SyntheticDefect}. While effective in some settings, training or fine-tuning generative models for a new inspection domain can require specialized expertise, compute resources, and, critically, sufficient domain data to learn realistic defect distributions.

Motivated by recent progress in multimodal large language models (MLLMs), we investigate whether off-the-shelf MLLMs can serve as low-barrier image generators for defect augmentation. These models can synthesize images from natural-language prompts and visual references without task-specific training, leveraging broad visual knowledge from large-scale pretraining \cite{Gemini2023}. In this work, we use an off-the-shelf MLLM image generator (e.g., Gemini 3 Pro Image) to generate synthetic insulator defect images conditioned on reference images.

This direction is particularly appealing in data-scarce inspection settings because, unlike training or fine-tuning domain-specific image generators, MLLM-based generation does not require collecting a large domain-specific corpus before producing usable samples. However, naive prompting of an MLLM with a single reference image often yields synthetic data that is not suitable for training: (i) diversity can be low, producing near-duplicates or limited coverage of defect appearance modes, and (ii) realism and label fidelity can vary substantially, introducing artifacts or ambiguous defect cues. These issues motivate a principled generation-and-filtering pipeline that targets both diversity and quality.

We propose a systematic framework for MLLM-based synthetic defect generation and demonstrate its effectiveness for ceramic insulator defect classification. We focus on two defect types with distinct visual signatures: \textit{shell} damage (chips and cracks at the rim) and \textit{glaze} damage (surface discoloration and degradation). Using group-based splitting and evaluating in a realistic low-data regime (10\% of the training split; 104 real training images, 152 validation images, and 308 test images), augmenting the training set with our synthetic images improves the classification F1 score from \FOneTenPct{} to \FOneSynthThreeX{} (20\% relative improvement), corresponding to a 4--5$\times$ data-efficiency gain.

Our main contributions are as follows:
\begin{enumerate}
  \item \textbf{Dual-reference conditioning for diversity:} we propose conditioning each synthetic image on two randomly sampled reference images from the same defect class to increase diversity and reduce single-reference mode collapse.

  \item \textbf{Prompt tuning and human verification for quality:} we introduce an iterative, class-specific prompt tuning procedure and a lightweight human-in-the-loop verification step to improve realism and label fidelity.

  \item \textbf{Embedding-based synthetic sample selection:} we propose an embedding-based filtering approach that selects high-quality synthetic images closest to their class centroids, improving the utility of generated data for downstream training.
\end{enumerate}

The remainder of this paper is organized as follows. Section~\ref{sec:related} reviews related work on power infrastructure inspection, defect data augmentation and synthesis, and foundation models for industrial vision. Section~\ref{sec:method} presents the proposed generation and filtering pipeline. Section~\ref{sec:experiments} reports experimental results and ablations, including empirical comparisons with RandAugment, zero-shot CLIP, and DreamBooth-based fine-tuning. Section~\ref{sec:discussion} discusses practical considerations and limitations, and Section~\ref{sec:conclusion} concludes the paper.

\section{Related Work}
\label{sec:related}

Our work intersects three research areas: deep learning for power infrastructure inspection, synthetic data augmentation for defect detection, and the emerging use of foundation models for industrial vision tasks.

\subsection{Deep Learning for Power Infrastructure Inspection}

The application of deep learning to power line inspection has grown rapidly with the proliferation of UAV-based imaging systems. Nguyen et al. \cite{Nguyen2018UAVReview} provided an early review of vision-based power line inspection, identifying deep learning as a promising direction while noting the challenge of limited training data. More recently, Liu et al. \cite{Liu2023InsulatorSurvey} surveyed deep learning methods specifically for insulator defect detection, cataloging approaches based on one-stage detectors (YOLO variants), two-stage detectors (Faster R-CNN), and semantic segmentation architectures.

Object detection frameworks dominate the literature. Liao et al. \cite{Liao2019StudyOP} applied an improved Faster R-CNN with soft non-maximum suppression to detect insulator defects in complex backgrounds, achieving high accuracy on their proprietary dataset. YOLO-based one-stage detectors are widely used in insulator inspection and can provide fast (sometimes real-time) inference, which is valuable for processing large volumes of inspection imagery \cite{Liu2023InsulatorSurvey,Redmon2016YOLO}.

Beyond detection, inspection workflows often require \emph{defect recognition}, i.e., assigning defect types or condition labels that can drive maintenance decisions. However, fine-grained defect recognition tends to be underrepresented relative to detection, partly because it requires curated, class-consistent labels and sufficient intra-class diversity. Public UAV inspection datasets do exist (e.g., InsPLAD \cite{InsPLAD2023} and Insulator Defect Detection \cite{defect_insulator_dataset}), but many studies still rely on proprietary collections and report results on limited defect examples \cite{Nguyen2018UAVReview,Liu2023InsulatorSurvey}.

Despite these advances, a consistent theme across the literature is \textbf{data scarcity}. Defect images are inherently rare, and utility companies often treat inspection data as proprietary, limiting the availability of widely adopted benchmarks. As a result, reported performance frequently depends on private datasets with relatively small numbers of labeled defect instances \cite{Nguyen2018UAVReview,Liu2023InsulatorSurvey}.

Existing deep-learning pipelines for power infrastructure inspection are increasingly mature, but limited labeled defect diversity (especially for defect-type recognition) remains a key barrier to robust generalization, motivating data-efficient augmentation methods.

\subsection{Data Augmentation for Defect Detection}

Data augmentation has long been recognized as essential for training robust vision models, particularly in data-scarce domains. Shorten and Khoshgoftaar \cite{Shorten2019DataAugSurvey} provide a comprehensive survey of augmentation techniques, categorizing them into geometric transformations (flipping, rotation, cropping), photometric transformations (color jitter, brightness adjustment), and more advanced methods including neural style transfer and GAN-based synthesis.

For defect detection specifically, traditional augmentations have limited utility. Geometric transformations cannot synthesize new defect patterns, i.e., they merely present existing defects from different viewpoints. This limitation has motivated the exploration of generative approaches.

\textbf{GAN-based augmentation} has shown promise in manufacturing contexts. Zhang et al. \cite{zhang2021defect} proposed Defect-GAN, which synthesizes defect samples by simulating defacement and restoration and composing defect foregrounds onto normal backgrounds to preserve background appearance. They report strong defect synthesis fidelity and show that the generated defects can improve downstream defect inspection performance in their experimental setting. Jain et al. \cite{Jain2022SyntheticDefect} applied GAN-based augmentation to a steel surface defect dataset, demonstrating that synthetic defects could improve classifier performance when real defect examples were limited.

However, GAN-based methods face two practical challenges in our setting. First, learning realistic defect distributions often benefits from substantially more defect examples than are available in inspection deployments, which undercuts the motivation for augmentation in data-scarce regimes. Second, GAN pipelines can be sensitive to training dynamics and hyperparameter choices, increasing implementation overhead in industrial settings.

\textbf{Diffusion-based synthesis} has recently become a strong alternative to GANs for image generation. Latent Diffusion Models \cite{Rombach2022LDM} and their descendants enable high-fidelity text-conditioned synthesis, and several adaptation techniques can personalize or control generation with limited data. For example, DreamBooth \cite{Ruiz2023DreamBooth} fine-tunes text-to-image diffusion models with only a few reference images, and ControlNet \cite{Zhang2023ControlNet} adds structural conditioning (e.g., edges or poses) to guide synthesis. While powerful, these approaches typically require access to model weights, careful fine-tuning/configuration, and additional engineering effort to ensure label fidelity for subtle defect categories.

Classical augmentation is limited to nuisance transformations, and many generative augmentation methods still require training or fine-tuning. This motivates training-free, low-barrier generation approaches with explicit quality control for defect label fidelity and realism.

\subsection{Foundation Models for Industrial Vision}

The emergence of large-scale foundation models that are pretrained on internet-scale datasets offers a new paradigm for data-efficient industrial vision. Rather than training task-specific models from scratch, practitioners can leverage pretrained representation models (e.g., CLIP \cite{Radford2021CLIP}) for recognition, and pretrained generative models (e.g., diffusion models \cite{Rombach2022LDM}) for synthesis.

In the industrial anomaly detection domain, several recent works exploit vision-language models. Gu et al. \cite{Gu2023AnomalyGPT} introduced AnomalyGPT, which uses a large vision-language model to detect and describe anomalies in few-shot settings by framing anomaly detection as a vision-language interaction task. Zhou et al. \cite{Zhou2024AnomalyCLIP} proposed AnomalyCLIP, which learns object-agnostic prompts for zero-shot anomaly detection. While these approaches show promise for detecting \emph{whether} an anomaly is present, fine-grained \emph{defect-type recognition} can be more sensitive to domain shift and subtle visual cues. Accordingly, we include CLIP-based zero-shot classification as a practical baseline in Section~\ref{sec:experiments}.

For synthetic data generation, foundation models offer zero-shot image synthesis through natural language prompts. Sapkota et al. \cite{Sapkota2025ImprovedYW} recently demonstrated this approach in agriculture, using LLM-generated synthetic images to augment apple detection training data. Their results showed that synthetic images from foundation models could improve detector performance, particularly when real training data was limited.

Foundation models enable both zero-shot recognition and powerful synthesis, but there remains a need for a practical, training-free defect augmentation pipeline that achieves diversity and label fidelity when generation is performed via prompting without assuming access to model weights or fine-tuning.

\section{Method}
\label{sec:method}

We present a practical synthetic augmentation pipeline designed for data-scarce defect recognition in a prompt-only image generation setting. The pipeline consists of: (i) dual-reference conditioning to increase diversity, (ii) iterative prompt refinement with human verification to ensure label fidelity, and (iii) embedding-based selection to automatically filter for the most in-distribution synthetic images before classifier training.

\subsection{Dataset and Preprocessing}

We utilize a public dataset of insulator images collected by UAVs during routine power line inspections \cite{defect_insulator_dataset}. The dataset contains two primary defect classes:
\begin{itemize}
    \item \textbf{Shell damage (Broken insulator shell):} As defined in the dataset \cite{defect_insulator_dataset}, physical breakage of the porcelain, i.e., chips or cracks at the rim or edge where material has broken off.
    \item \textbf{Glaze damage (Flashover damage):} As defined in the dataset \cite{defect_insulator_dataset}, insulator shells labeled with flashover damage, i.e., surface discoloration of the ceramic glaze (matte, faded, or visible color change), often with a lighter edge around affected areas.
\end{itemize}

We begin from the raw UAV dataset \cite{defect_insulator_dataset}, which provides insulator bounding boxes and defect-type annotations. For each image, we compute a single tight crop by taking the minimum enclosing rectangle over all annotated insulator bounding boxes in that image. During training, we apply a random crop expansion to include context, then pad the crop to a square and resize to $224 \times 224$ pixels for classifier training. For synthetic generation, we use the cropped images as reference images to the image generator.

Next, we curate labels to obtain an unambiguous defect-type recognition task. Although the raw annotations can occasionally include multiple defect types on the same physical insulator, such cases are rare in this dataset; we exclude the few dual-defect (shell+glaze) insulator groups during preprocessing. We then retain only \emph{class-consistent} groups labeled as shell-only or glaze-only, so each cropped sample used in our experiments has a one-hot ground-truth label (\textit{shell} \emph{or} \textit{glaze}).

We then split the curated dataset by \textit{insulator groups} into 70\% train, 10\% validation, and 20\% test, ensuring that each physical insulator string appears in only one split and preventing leakage across viewpoints of the same equipment. Here, an \textit{insulator group} denotes multiple images of the same physical insulator captured from different viewpoints. To study data scarcity, we additionally run a real-only sweep over multiple fractions of the \emph{training} split (e.g., 5\%, 10\%, 20\%, \ldots, 100\%). These fractions are defined over \emph{insulator groups} (not individual images) and are constructed cumulatively, so smaller-fraction training sets are strict subsets of larger ones (e.g., the 10\% training set is a subset of the 20\% training set); this preserves group-based splitting at every fraction and enables fair comparisons across data regimes.

Finally, we adopt a defect-only scope: we do not include an explicit normal class in any split. We train a two-output classifier with binary cross-entropy (one logit per defect type), which does not impose mutual exclusivity even though the curated labels are one-hot. In practical deployments, normal (healthy) insulator images are abundant, so there is typically no need to augment the normal class with synthetic images; synthetic augmentation is most valuable for the rare defect classes.

\subsection{Iterative Synthetic Image Generation}

\begin{figure}[!htb]
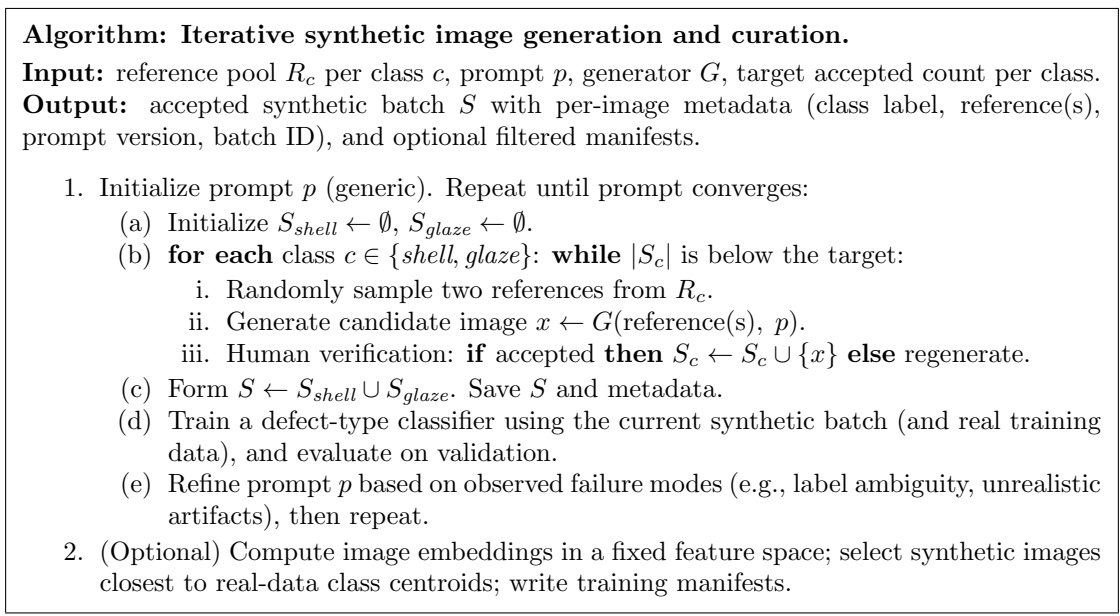

\centering
\footnotesize
\setlength{\fboxsep}{6pt}
\fbox{\begin{minipage}{0.95\linewidth}
\textbf{Algorithm: Iterative synthetic image generation and curation.}\vspace{0.25em}\\
\textbf{Input:} reference pool \(R_c\) per class \(c\), prompt \(p\), generator \(G\), target accepted count per class.\\
\textbf{Output:} accepted synthetic batch \(S\) with per-image metadata (class label, reference(s), prompt version, batch ID), and optional filtered manifests.
\begin{enumerate}
    \item Initialize prompt \(p\) (generic). Repeat until prompt converges:
    \begin{enumerate}
        \item Initialize \(S_{\textit{shell}} \leftarrow \emptyset\), \(S_{\textit{glaze}} \leftarrow \emptyset\).
        \item \textbf{for each} class \(c\in\{\textit{shell},\textit{glaze}\}\): \textbf{while} \(|S_c|\) is below the target:
        \begin{enumerate}
            \item Randomly sample two references from \(R_c\).
            \item Generate candidate image \(x \leftarrow G(\text{reference(s)},\;p)\).
            \item Human verification: \textbf{if} accepted \textbf{then} \(S_c \leftarrow S_c \cup \{x\}\) \textbf{else} regenerate.
        \end{enumerate}
        \item Form \(S \leftarrow S_{\textit{shell}} \cup S_{\textit{glaze}}\). Save \(S\) and metadata.
        \item Train a defect-type classifier using the current synthetic batch (and real training data), and evaluate on validation.
        \item Refine prompt \(p\) based on observed failure modes (e.g., label ambiguity, unrealistic artifacts), then repeat.
    \end{enumerate}
    \item (Optional) Compute image embeddings in a fixed feature space; select synthetic images closest to real-data class centroids; write training manifests.
\end{enumerate}
\end{minipage}}
\caption{Pseudocode for the proposed iterative generation, verification, and prompt-refinement loop.}
\label{alg:iterative_generation}
\end{figure}

Our pipeline generates high-fidelity synthetic defect images through iterative prompt refinement and human verification, as shown in Figure~\ref{alg:iterative_generation}.

\subsubsection{Model Selection}
We utilize \textbf{Gemini 3 Pro Image} for image generation. This model was selected after preliminary testing showed that other models struggled to maintain geometric fidelity for industrial inspection tasks, often generating unrealistic insulators or unrealistic defect patterns.

\subsubsection{Dual-Reference Generation Strategy}

Single-reference generation suffers from mode collapse: Synthetic images are too similar to their source reference. To address this, we introduce a \textbf{dual-reference strategy} where each generation uses two randomly sampled reference images of the same defect class.

The prompt instructs the model to:
\begin{enumerate}
    \item Study both reference images for defect characteristics
    \item Generate a new image combining structural elements from both
    \item Vary insulator color, background, angle, and lighting
    \item Ensure defect visibility matches reference scale
\end{enumerate}

\textbf{Batching and label scope.} Dual-reference generation produces single-label synthetic images (\textit{shell} \emph{or} \textit{glaze}) by pairing references from the same class. To reflect a realistic setting, we use the 10\% training fraction as the reference image pool. This reference pool contains 52 images per class. To better monitor generation quality, we generate images in batches; each batch contains 104 synthetic images (52 per class).

\textbf{Diversity ratio.} We quantify diversity of the images in a fixed feature space using an ImageNet-pretrained ResNet-18 embedding. Let \(d_{\text{syn}\rightarrow\text{ref}}\) be the mean Euclidean distance between each synthetic image and its reference image; for dual-reference generations, we compute the distance to \emph{both} references and take the minimum (i.e., the closest-reference distance). Let \(d_{\text{real-pair}}\) be the mean Euclidean distance of randomly sampled pairs of real images. We define the diversity ratio as \(d_{\text{syn}\rightarrow\text{ref}} / d_{\text{real-pair}}\). When this ratio approaches 0, synthetic images are near-duplicates of their references; when it is close to 1, the synthetic-to-reference distance matches typical real-pair variation, which we treat as a desirable diversity target; ratios much larger than 1 indicate synthetic images that are farther from references than typical real images are from each other.

\begin{figure}[!htb]
    \centering
    \includegraphics[width=1\textwidth]{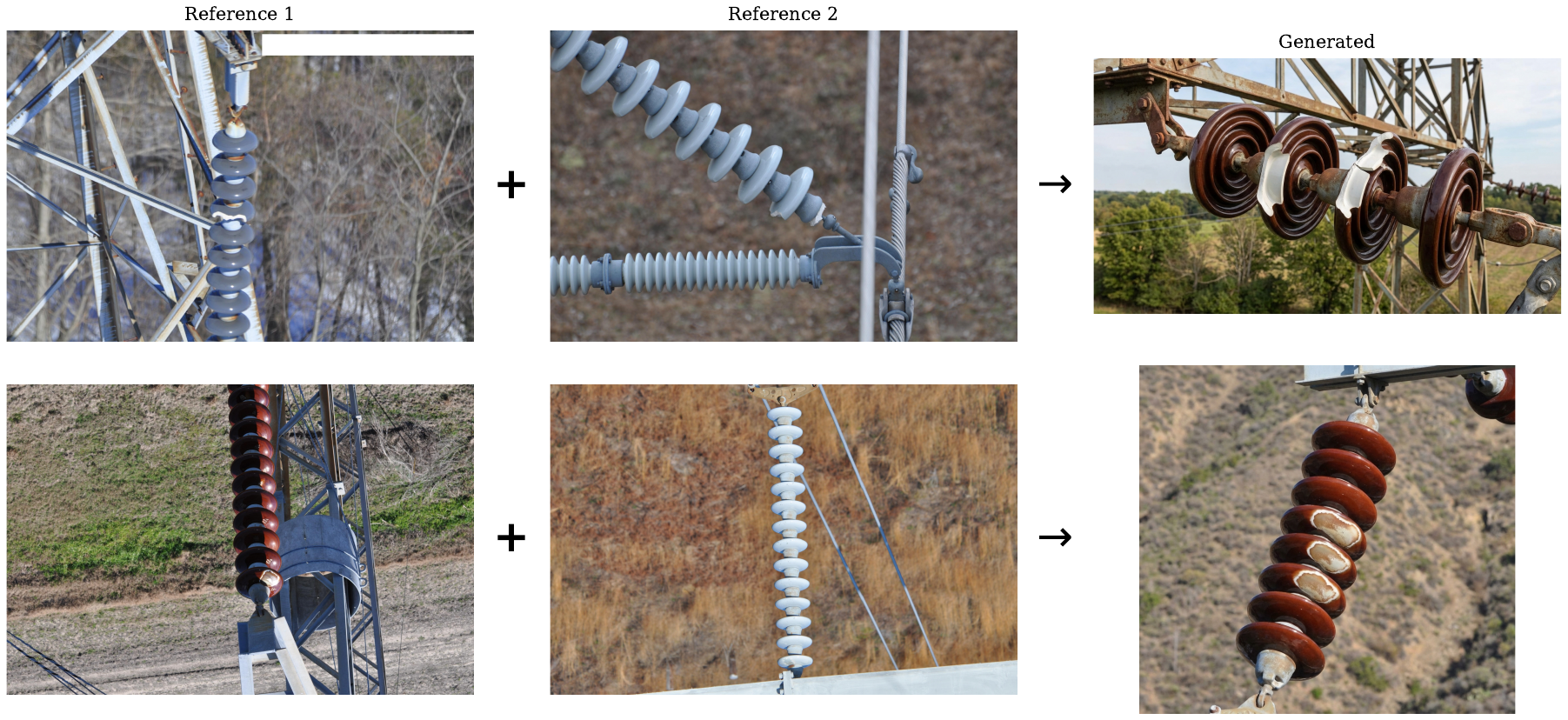}
    \caption{Dual-reference generation demonstration.}
    \label{fig:conceptual_demo}
    \end{figure}

The dual-reference generation strategy is demonstrated in Figure~\ref{fig:conceptual_demo}.

\subsubsection{Class-Specific Prompt Engineering}

Naively using a generic prompt often yields label-ambiguous or physically implausible defects (e.g., missing class-defining cues, mixing multiple defect mechanisms, or producing unrealistic severity). We therefore iteratively refined class-specific prompts following the failure-driven loop in Figure~\ref{alg:iterative_generation}. Prompt refinement followed four general principles:
\begin{itemize}
    \item \textbf{Define the class in positive terms:} describe what the defect \emph{is} (appearance, location on the component, and material/texture cues) in a way that a human annotator would recognize.
    \item \textbf{Add disambiguating cues:} enforce at least one diagnostic visual signature that separates the class from visually similar failure modes.
    \item \textbf{Bound scale and frequency:} specify the acceptable range of defect extent (relative to the object scale and/or the references) and constrain how many distinct affected regions appear in a single image.
    \item \textbf{Specify negative constraints and allowed variation:} explicitly prohibit common generation errors (wrong object/material, wrong defect type, implausible geometry/texture) while allowing nuisance variation in viewpoint, lighting, and background to increase diversity.
\end{itemize}

\subsubsection{Human-in-the-Loop Verification}

Even with prompt refinement, some images may still be unrealistic or out-of-distribution. Such images can be easily identified by human annotators such as domain specialists; in our experiments, verification was performed by an annotator familiar with the defect types. To ensure the quality of the generated images, we include a human-in-the-loop verification step before feeding the images to the model for training. Note that the goal of this step is not to find the best images, but to ensure the obvious failures are caught.

\begin{figure}[!htb]
    \centering
    \includegraphics[width=1\textwidth]{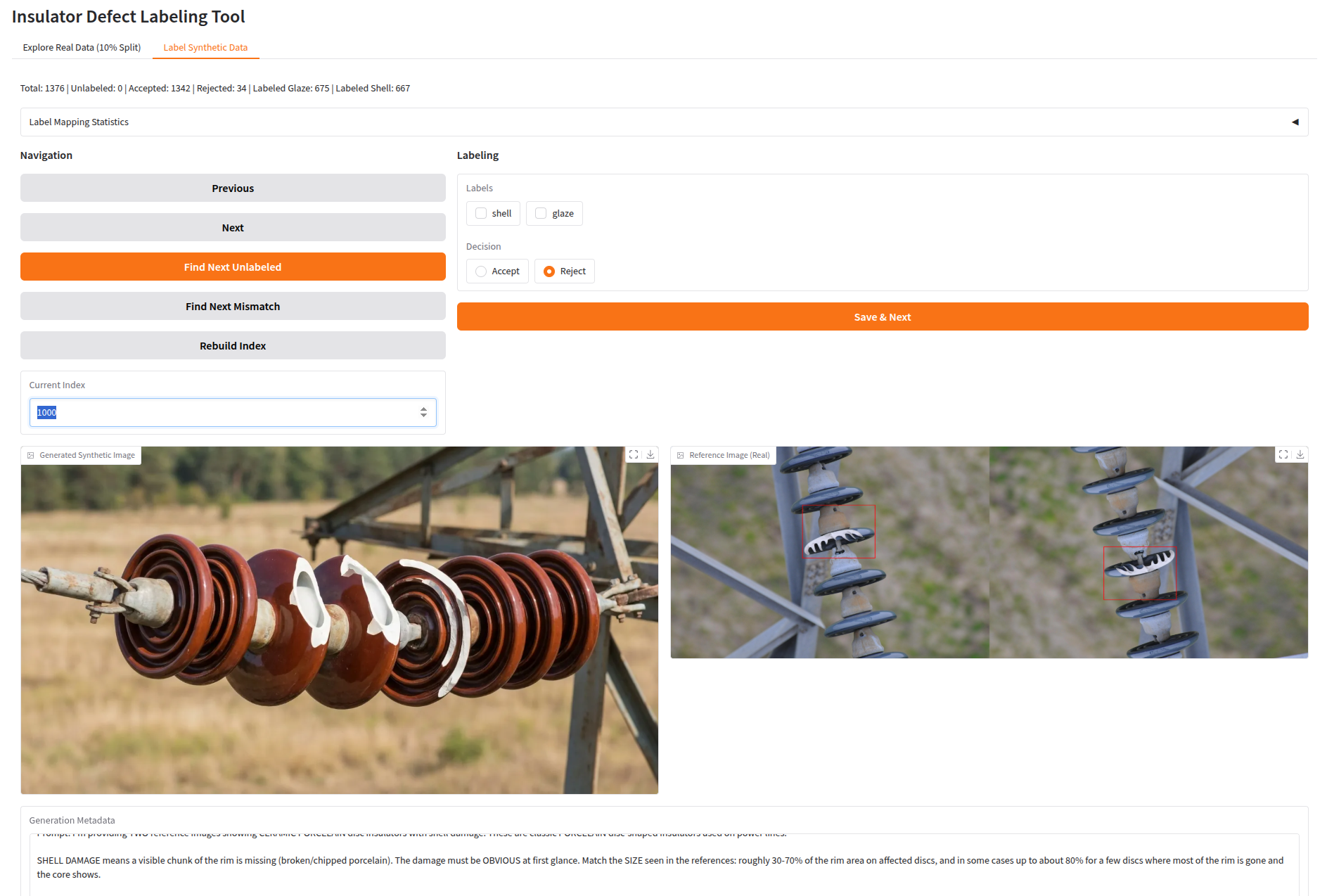}
    \caption{Human-in-the-loop verification interface.}
    \label{fig:human_verification_ui}
\end{figure}

We built a custom interface for human verification, as shown in Figure~\ref{fig:human_verification_ui}. The interface allows the annotator to explore the real data and learn the distribution of the real defects. The annotator can then switch to the labeling page. On the labeling page, the generated sample is shown along with the reference image(s). The annotator can then accept or reject the generated sample.

In our experiments, to ensure each batch contains the same number of images as the 10\% training fraction, we re-generate the rejected samples until the batch reaches the target number of accepted images per class. As shown later, the rejection rate is quite low, so in practice the regeneration effort is trivial for this task.

\subsubsection{Embedding-Based Quality Filtering}

While human verification catches obvious failures, subtle quality variations remain, e.g., images may be technically correct but exhibit varying levels of realism, defect clarity, or class-discriminative features. We introduce an automated filtering step based on embedding-space class consistency.

The goal of this step is to select the images that are most likely to be in the true distribution of the real data. The core idea is to select the images that are closest to the class centroid in the high-dimensional embedding space. For each defect class, we compute the centroid of the real training images (10\% training fraction) in this embedding space. We then rank all synthetic images by their Euclidean distance to their class centroid, i.e., images closer to the centroid are more clearly in the correct class region of the feature space and more likely to share visual characteristics with real defect images.

To ensure the robustness of the embedding space, we choose to use the ResNet backbone pretrained on ImageNet instead of the model fine-tuned on the 10\% training fraction. The pretrained ImageNet features, trained on millions of diverse images, provide a more reliable embedding space for quality assessment. We extract 512-dimensional feature embeddings from the layer before the classification head. A later section shows that image quality varies from batch to batch, but the embedding-based selection provides consistent and scalable improvements in model performance.

\subsection{Classification Framework}

\textbf{Architecture:} ResNet-18 pretrained on ImageNet, with the final fully-connected layer replaced to output one logit per defect type (shell, glaze).

\textbf{Training Protocol:}
\begin{itemize}
    \item Optimizer: AdamW with learning rate $3 \times 10^{-4}$, weight decay $10^{-4}$
    \item Batch size: 128
    \item Epochs: 20 in total, using the model with the best F1 score on the validation set for testing.
    \item Loss: Binary cross-entropy
    \item Augmentation: Horizontal flip only (minimal augmentation to isolate synthetic data effect)
\end{itemize}

\textbf{Loss Function Design:} Our curated dataset assigns a single defect type per image after removing dual-defect groups during preprocessing. Nevertheless, real components may exhibit multiple concurrent defects, and we therefore use binary cross-entropy (BCE) rather than softmax cross-entropy. BCE treats each defect type as an independent binary prediction, preventing the model from exploiting a mutual-exclusivity shortcut (i.e., if not shell, then glaze) and yielding a classifier that can predict multiple defect types on the same component. At inference time, we output a separate sigmoid probability for each defect type.

\textbf{Mixed Training:} When training with synthetic data, we concatenate the real training set (10\% training fraction) with synthetic images. To align with the random crop expansion on the real training images, we apply a random zoom-out augmentation to the synthetic images. There is no augmentation on the validation and test sets.

\section{Experiments and Results}
\label{sec:experiments}

\subsection{Experimental Setup}

\textbf{Dataset and model:} We follow the dataset definition, group-based splitting, preprocessing, and classifier training protocol described in Section~\ref{sec:method}.

\textbf{Metrics:} We report F1-score on the two defect labels. For supervised models, we apply a 0.5 threshold to the sigmoid outputs and compute precision/recall/F1 by aggregating true positives, false positives, and false negatives across the two labels (equivalently, a micro-averaged F1 over label decisions). We report mean $\pm$ standard deviation over 3 random seeds.

\subsection{Baselines}

\textbf{Zero-shot CLIP.} To assess whether general-purpose vision-language models can solve this task without any domain-specific training, we evaluate CLIP ViT-L/14 \cite{Radford2021CLIP} in a zero-shot setting. For each test image, we compute CLIP image-text similarity scores for two text prompts (\textit{a photo of a ceramic insulator with shell damage} and \textit{a photo of a ceramic insulator with glaze damage}) and predict the label corresponding to the higher-scoring prompt.

\textbf{RandAugment.} We evaluate RandAugment \cite{Cubuk2020RandAugment} as a strong traditional augmentation baseline applied to supervised training on real images.

\textbf{DreamBooth.} As a standard diffusion baseline, we also evaluate DreamBooth fine-tuning \cite{Ruiz2023DreamBooth} of Stable Diffusion 2.1 for defect synthesis, then train the same downstream classifier on the resulting synthetic data.

\subsection{Impact of Training Data Size and RandAugment}

We first run a real-only fraction sweep (5\%--100\%) to quantify the severity of data scarcity in this task. Performance increases with data volume but exhibits diminishing returns; full numeric results (baseline F1 and image counts) are reported in Appendix Table~\ref{tab:randaugment_comparison}.

At 10\% of the training split (our target scenario), the model achieves only \FOneTenPct{} F1, compared to \FOneFull{} F1 with full training data, a gap of \FOneGapFullMinusTen{} that motivates synthetic augmentation. The full training set contains on the order of 500 images per class; a central goal of this work is to improve defect-type recognition when such large per-class counts are not available. In practice, collecting defect imagery is time-consuming (inspections are periodic, and defective components are relatively rare), and realistic data-collection targets for a given defect type are often on the order of tens to low hundreds of images. We therefore adopt the 10\% fraction (52 images per class) as our primary low-data regime to align with this practical setting.

\begin{figure}[!htb]
\centering
\includegraphics[width=0.7\textwidth]{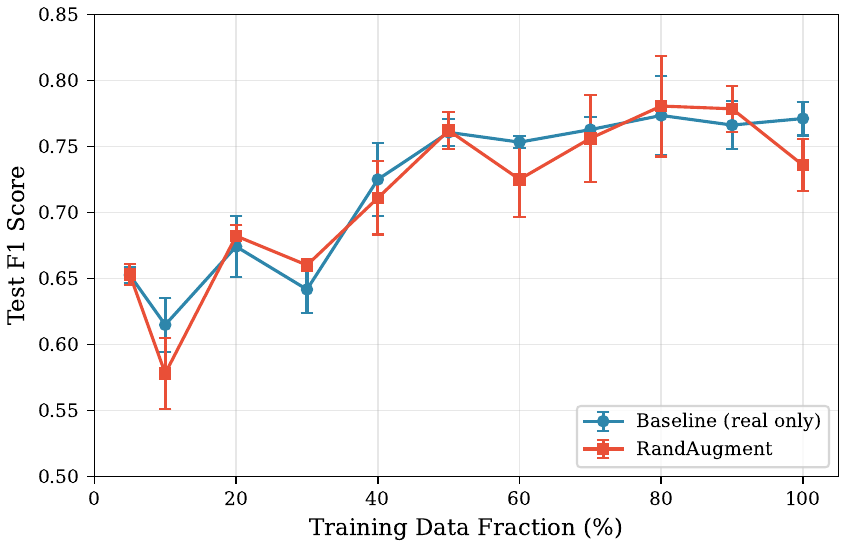}
\caption{Baseline vs.\ RandAugment across training fractions. RandAugment shows inconsistent behavior and does not reliably improve defect-type recognition in the data-scarce regime.}
\label{fig:randaugment_sweep}
\end{figure}

\textbf{Traditional augmentation across fractions.} We also evaluate RandAugment across the same fraction sweep to test whether classical geometric/photometric augmentation helps in this fine-grained defect setting. Figure~\ref{fig:randaugment_sweep} summarizes the trend: RandAugment yields no consistent improvements in the low-data regime and can sometimes even degrade performance. Full numeric results are reported in Appendix Table~\ref{tab:randaugment_comparison}.

These results suggest the bottleneck in defect-type recognition is not geometric or photometric invariance, but \textit{semantic diversity}: the classifier needs exposure to qualitatively different crack patterns, discoloration extents, and defect locations. RandAugment perturbs pixels without creating new defect manifestations, so it cannot reliably address scarcity in this setting. This motivates the proposed \textbf{conceptual synthesis} approach: conditioned on real reference images and explicit prompt constraints, an MLLM can generate physically plausible, semantically novel defect instances rather than transformed copies of existing ones.

\subsection{Ablation Studies}

\subsubsection{Prompt Engineering Impact}

We tested two prompt versions for dual-reference generation. Prompt V1 is a basic damage description, while prompt V2 enhanced specifications like 30--70\% rim damage, white edge detection, and fracture details. Detailed V1 and V2 prompts are reported in Appendix Section~\ref{sec:appendix_prompts}. V2 prompts yield substantially better downstream performance than V1 despite similar diversity ratios, demonstrating that prompt quality affects generation fidelity.

\begin{table}[!htb]
\centering
\caption{Dual-Reference Prompt Refinement (V1 $\rightarrow$ V2)}
\label{tab:prompt_refinement}
\begin{tabular}{lccc}
\hline
\textbf{Prompt Version} & \textbf{Batches} & \textbf{Diversity Ratio} & \textbf{Test F1} \\
\hline
Dual-ref V1 & 0,1,2 & 1.06 & 0.640 $\pm$ 0.021 \\
Dual-ref V2 & 3,4,5 & 1.09 & \textbf{0.700 $\pm$ 0.039} \\
\hline
\end{tabular}
\end{table}

\subsubsection{Single-Reference vs. Dual-Reference Generation}

We compare single-reference generation and the proposed dual-reference strategy under V2-style prompts. Dual-reference increases embedding-space diversity and improves downstream test performance. \textbf{Diversity Ratio} measures how far synthetic images are from their reference(s) compared to real image pairwise distances. Single-reference generation produces images too similar to references, while dual-reference achieves ratio $>$1.0, indicating true diversity. Under V2 prompts, the dual-reference strategy improves test F1 by +0.048 compared to single-reference.

\begin{table}[!htb]
\centering
\caption{Single-Reference vs. Dual-Reference Generation (V2 Prompts)}
\label{tab:single_vs_dual}
\begin{tabular}{lccc}
\hline
\textbf{Strategy} & \textbf{Diversity Ratio} & \textbf{Test F1} \\
\hline
Single-ref (V2; 3 batches) & 0.68 & 0.652 $\pm$ 0.030 \\
Dual-ref (V2; batches 3,4,5) & 1.09 & \textbf{0.700 $\pm$ 0.039} \\
\hline
\end{tabular}
\end{table}

\subsubsection{Batch Quality Variation}

\begin{figure}[!htb]
\centering
\includegraphics[width=0.85\textwidth]{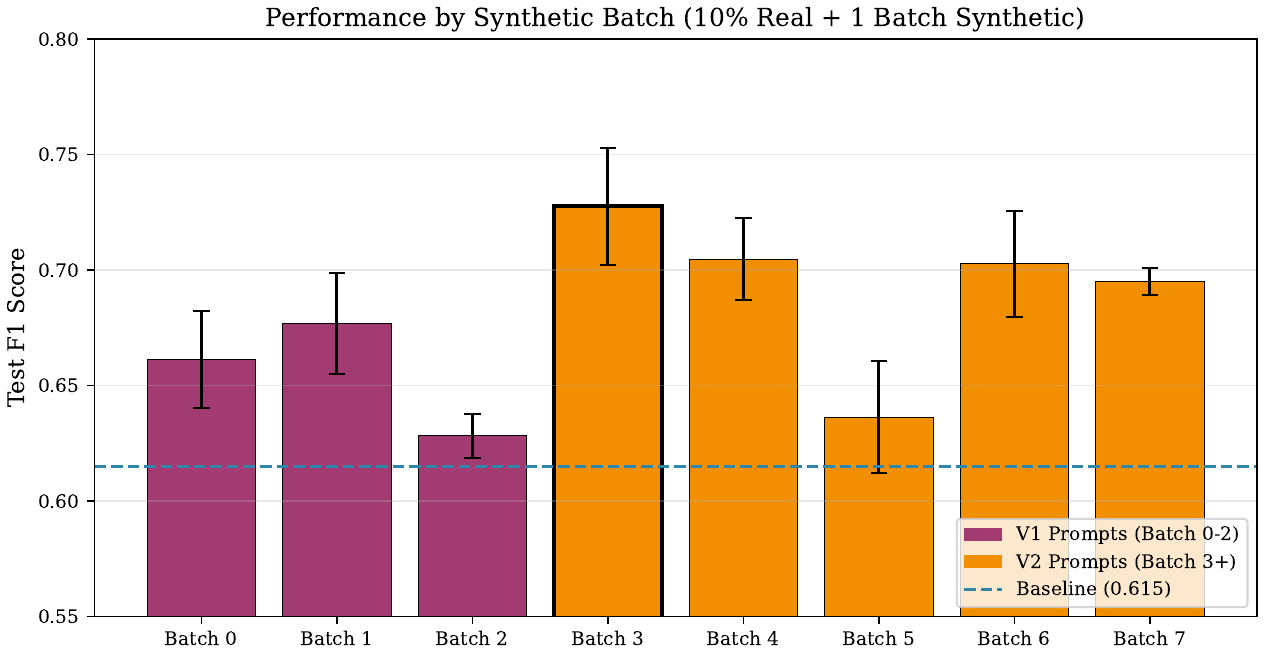}
\caption{Test F1 performance by synthetic batch for dual-reference generations. V1 prompt batches (0--2) shown in purple, V2 prompt batches (3--7) shown in orange. Dashed line indicates 10\% baseline (\FOneTenPct{}).}
\label{fig:batch_comparison}
\end{figure}

In dual-reference settings, we randomly sample the pair of reference images without replacement. Even with the same settings, the performance variation across synthetic batches is significant as shown in Figure~\ref{fig:batch_comparison}. Batches 0, 1, and 2 use prompt V1, while batches 3--7 use prompt V2. Among V2 prompt batches, batch 3 achieves 0.728 F1 while batch 5 achieves only 0.636 F1. This 0.092 F1 gap within the same generation pipeline highlights the importance of quality filtering.

\subsubsection{Quality vs. Quantity Trade-off and Selection Scalability}

Counter-intuitively, adding more synthetic batches does not always improve performance. However, principled embedding-based selection demonstrates consistent improvements across multiple scales. As can be seen in Table~\ref{tab:quantity}, manual batch selection exhibits non-monotonic behavior: combining batches often degrades performance compared to the best single batch (batch 3). In contrast, \textbf{embedding-based selection yields strong improvements across multiple scales}. In our main protocol, embedding-based selection ranks candidates from the full dual-reference synthetic pool (batches 0--7; 832 accepted images) by distance to class centroids computed from the 10\% real \emph{training} split only, and then selects the top $N$ per class for training. As more images are selected per class, the selection threshold becomes less strict (images farther from the real centroid are included), which can increase both the average performance and the run-to-run variance.

\begin{table}[!ht]
\centering
\caption{Effect of Synthetic Data Quantity and Selection Strategy}
\label{tab:quantity}
\begin{tabular}{lccc}
\hline
\textbf{Configuration} & \textbf{Selection} & \textbf{Synthetic Images} & \textbf{Test F1} \\
\hline
\multicolumn{4}{c}{\textit{Manual batch selection (unreliable)}} \\
\hline
Batch 3 only & Manual & 104 & 0.728 $\pm$ 0.025 \\
Batches 3,4 & Manual & 208 & 0.676 $\pm$ 0.029 \\
Batches 3,4,5 & Manual & 312 & 0.700 $\pm$ 0.039 \\
\hline
\multicolumn{4}{c}{\textit{Embedding-based selection (reliable, scalable)}} \\
\hline
Top 52/class (1$\times$) & Embedding & 104 & 0.693 $\pm$ 0.000 \\
Top 104/class (2$\times$) & Embedding & 208 & 0.705 $\pm$ 0.018 \\
Top 156/class (3$\times$) & Embedding & 312 & \textbf{0.739 $\pm$ 0.035} \\
\hline
\end{tabular}
\end{table}

This scalability analysis demonstrates that the embedding-based method is \textit{robust and predictable} in that it avoids manual batch cherry-picking and provides consistent gains. The 3$\times$ configuration achieves the best mean performance in our experiments, while exhibiting higher variance, consistent with a quality-quantity trade-off as more borderline synthetic samples are admitted.

\subsection{Efficacy of Synthetic Augmentation}

We now compare our final synthetic augmentation pipeline against representative baselines in the target low-data regime (10\% real training split). For all supervised settings, we train the same ResNet-18 classifier described in Section~\ref{sec:method} and report mean $\pm$ standard deviation over 3 random seeds.

Table~\ref{tab:main_comparison} compares five settings: (i) \textit{Real-only (10\%)} supervised training (baseline), (ii) \textit{RandAugment} on the same 10\% split, (iii) \textit{DreamBooth} synthetic augmentation, (iv) the \textit{Proposed method} (dual-reference + human verification + embedding-based selection; 3$\times$ synthetic mixed with 10\% real), and (v) \textit{CLIP (zero-shot)} with no supervised training (two text prompts; higher similarity wins).

For DreamBooth, we fine-tune Stable Diffusion 2.1 on all 52 images per class from the 10\% training split, generate 104 synthetic images per class, and then train the same downstream classifier on the synthetic+real mix.

\begin{table}[!ht]
\centering
\caption{Comparison with Representative Baselines (10\% Real Training Data)}
\label{tab:main_comparison}
\begin{tabular}{lcc}
\hline
\textbf{Method} & \textbf{Test F1} & \textbf{$\Delta$ vs Baseline} \\
\hline
CLIP (zero-shot) & 0.429 & -0.186 \\
\hline
Real-only (10\%) & 0.615 $\pm$ 0.020 & --- \\
RandAugment & 0.578 $\pm$ 0.027 & -0.037 \\
DreamBooth & 0.622 $\pm$ 0.021 & +0.007 \\
Proposed (10\% + selected 3$\times$ synthetic) & \textbf{0.739 $\pm$ 0.035} & \textbf{+0.124} \\
\hline
\end{tabular}
\end{table}

\textbf{Robustness to stronger backbones and frozen-feature baselines.} To assess whether the synthetic-data benefit is specific to a particular classifier, we repeated key experiments using a stronger supervised backbone (ResNet-50) and using frozen pretrained encoders with a linear probe (CLIP and DINOv2 \cite{Oquab2023DINOv2}). A \textit{linear probe} freezes the pretrained encoder and trains only a linear classifier on top of extracted features, isolating representation quality from end-to-end fine-tuning. In all cases, adding embedding-selected synthetic images improves performance (Table~\ref{tab:robustness_baselines}).

\begin{table}[!ht]
\centering
\caption{Robustness of Synthetic Augmentation Across Backbones and Frozen-Feature Baselines}
\label{tab:robustness_baselines}
\begin{tabular}{lcc}
\hline
\textbf{Model / Baseline} & \textbf{Training data} & \textbf{Test F1} \\
\hline
ResNet-18 (supervised) & 10\% real only & 0.615 $\pm$ 0.020 \\
ResNet-18 (supervised) & 10\% real + synthetic (3$\times$) & \textbf{0.739 $\pm$ 0.035} \\
\hline
ResNet-50 (supervised) & 10\% real only & 0.586 $\pm$ 0.021 \\
ResNet-50 (supervised) & 10\% real + synthetic (3$\times$) & 0.700 $\pm$ 0.003 \\
\hline
CLIP ViT-L/14 (linear probe) & 10\% real only & 0.601 \\
CLIP ViT-L/14 (linear probe) & 10\% real + synthetic (3$\times$) & 0.656 \\
DINOv2 (linear probe) & 10\% real only & 0.555 \\
DINOv2 (linear probe) & 10\% real + synthetic (3$\times$) & 0.675 \\
\hline
\end{tabular}
\end{table}

\subsection{Generation Cost and Rejection Analysis}

Table~\ref{tab:generation_cost} presents the practical costs of synthetic image generation using the Gemini 3 Pro Image API.

\begin{table}[!ht]
\centering
\caption{Generation Cost and Rejection Rates by Batch}
\label{tab:generation_cost}
\begin{tabular}{lccccc}
\hline
\textbf{Batch} & \textbf{Prompt} & \textbf{Requests} & \textbf{Rejected} & \textbf{Rejection \%} & \textbf{Cost (USD)} \\
\hline
0 & V1 & 104 & 0 & 0.0\% & \$14.14 \\
1 & V1 & 106 & 2 & 1.9\% & \$14.41 \\
2 & V1 & 107 & 3 & 2.9\% & \$14.54 \\
3 & V2 & 107 & 3 & 2.9\% & \$14.57 \\
4 & V2 & 112 & 8 & 7.7\% & \$15.25 \\
5 & V2 & 108 & 4 & 3.8\% & \$14.71 \\
6 & V2 & 106 & 2 & 1.9\% & \$14.44 \\
7 & V2 & 106 & 2 & 1.9\% & \$14.44 \\
\hline
\textbf{Total} & --- & 856 & 24 & 2.9\% & \$116.49 \\
\hline
\end{tabular}
\end{table}

We compute generation cost from per-request token usage recorded in metadata, using API pricing of \$2 per million input tokens, \$12 per million output text tokens, and \$120 per million output image tokens. Under this pricing, the total cost for our eight batches is \textbf{\$116.49} (Table~\ref{tab:generation_cost}), substantially less than a single UAV inspection flight.

Rejection rates remained low; in Table~\ref{tab:generation_cost}, rejection rate is defined as the number of rejected samples divided by the target accepted count per batch because the regenerated images are accepted without exception in our experiments. Most rejections were due to unrealistic defect appearance.

Human verification is lightweight: our interface presents batches of generated images; an annotator can review and flag rejections in 15--20 minutes per batch, spending approximately 10 seconds per image on binary accept/reject decisions.

These costs are small compared to the operational expense of UAV inspections or the cost of training and maintaining a custom generative model (GPU compute, engineering time). This economic analysis supports the practical viability of MLLM-based augmentation for industrial applications.

\section{Discussion}
\label{sec:discussion}

Our results suggest that \textbf{conceptual synthesis} with an off-the-shelf Multimodal Large Language Model (MLLM) is a practical alternative to training a custom generator in data-scarce inspection settings. Unlike pixel-level augmentation, MLLM generation can introduce semantically novel defect manifestations while preserving physical plausibility, and embedding-based selection makes gains predictable across batches. Off-the-shelf APIs are attractive because they require no GPU infrastructure and allow rapid prompt iteration; in our experiments, DreamBooth-style diffusion personalization yields only marginal gains relative to the proposed method \cite{Ruiz2023DreamBooth,Rombach2022LDM}.

In deployment, synthetic augmentation is most valuable for \textbf{rare defect classes}, while normal (healthy) images are typically abundant. A full inspection pipeline would combine a normal-vs-defect detector with a defect-type classifier, and extending to severity grading or concurrent defects would require additional labeled reference images and prompt development. Generating a larger synthetic pool can support stricter selection at a fixed acceptance threshold, improving predictability without relying on manual batch choice.

\textbf{Limitations.} \textit{Scope and generalization:} our results are obtained on a single public UAV inspection dataset \cite{defect_insulator_dataset} for one component type and two defect categories, and we did not evaluate cross-dataset transfer under distribution shift (different sensors, backgrounds, or component geometries). \textit{Synthetic artifacts:} despite dual-reference conditioning and human verification, MLLMs can produce subtle artifacts; our rejection analysis captures obvious failures, but accepted images may still introduce noise, motivating additional automated screening.

\section{Conclusion}
\label{sec:conclusion}

This paper addressed the challenge of training defect-type classifiers when defect images are scarce (i.e., a common situation in power infrastructure inspection). We presented a practical framework that leverages off-the-shelf Multimodal Large Language Models (MLLMs) to generate synthetic defect images without custom model training, combining (i) \textbf{dual-reference conditioning} to increase diversity, (ii) \textbf{iterative class-specific prompt engineering} with lightweight human verification to improve label fidelity, and (iii) \textbf{embedding-based selection} that ranks synthetic samples by distance to class centroids computed from the real \emph{training} split only. On ceramic insulator defect-type classification using a public UAV dataset \cite{defect_insulator_dataset} in a realistic 10\% training-data regime, embedding-selected synthetic augmentation improves test F1 from \FOneTenPct{} to \FOneSynthThreeX{} (20\% relative), corresponding to a 4--5$\times$ data-efficiency gain; the improvement persists under stronger supervised backbones and frozen-feature linear-probe baselines. Future work includes evaluating cross-dataset generalization under distribution shift, extending to multi-class inspection pipelines including normal insulators and severity grading, and exploring alternative embedding spaces for selection.

\appendix

\section{Supplementary: RandAugment Fraction Sweep Numbers}
\label{sec:appendix_randaugment}

\begin{table}[!ht]
\centering
\caption{RandAugment vs. Supervised Baseline Across Training Data Fractions}
\label{tab:randaugment_comparison}
\begin{tabular}{lcccc}
\hline
\textbf{Fraction} & \textbf{Images} & \textbf{Baseline F1} & \textbf{RandAugment F1} & \textbf{$\Delta$} \\
\hline
5\% & 48 & 0.653 $\pm$ 0.006 & 0.653 $\pm$ 0.008 & +0.000 \\
10\% & 104 & 0.615 $\pm$ 0.020 & 0.578 $\pm$ 0.027 & -0.037 \\
20\% & 212 & 0.674 $\pm$ 0.023 & 0.682 $\pm$ 0.008 & +0.008 \\
30\% & 316 & 0.642 $\pm$ 0.017 & 0.660 $\pm$ 0.004 & +0.018 \\
40\% & 428 & 0.725 $\pm$ 0.028 & 0.711 $\pm$ 0.028 & -0.014 \\
50\% & 532 & 0.761 $\pm$ 0.010 & 0.762 $\pm$ 0.014 & +0.001 \\
60\% & 640 & 0.753 $\pm$ 0.004 & 0.725 $\pm$ 0.028 & -0.028 \\
70\% & 748 & 0.763 $\pm$ 0.009 & 0.756 $\pm$ 0.033 & -0.007 \\
80\% & 856 & 0.774 $\pm$ 0.030 & 0.781 $\pm$ 0.038 & +0.007 \\
90\% & 960 & 0.766 $\pm$ 0.018 & 0.779 $\pm$ 0.017 & +0.012 \\
100\% & 1072 & 0.771 $\pm$ 0.013 & 0.736 $\pm$ 0.020 & -0.035 \\
\hline
\end{tabular}
\end{table}

\section{Complete Prompt Templates}
\label{sec:appendix_prompts}

For reproducibility, we provide the complete prompts used for synthetic image generation.

\subsection{Single-Reference Prompts (V2 Style)}

\textbf{Glaze Damage (Single-Reference):}

\begin{quote}
\small
Based on the provided reference image of a CERAMIC PORCELAIN disc insulator, generate a realistic variation for a dataset.

GLAZE DAMAGE is a COLOR CHANGE in the ceramic surface itself---areas where the shiny glaze has become matte, faded, or discolored. The damaged areas are FLUSH with the surface (not raised, not peeling).

CRITICAL: Real glaze damage shows a DISTINCT LIGHT/WHITE EDGE or border around the damaged patch. This looks like a thin white or very light-colored intermediate layer between the outer colored glaze (red/brown) and the inner white ceramic. You MUST include this white edge pattern around damage patches.

REQUIREMENTS:
\begin{itemize}
    \item MUST be ceramic/porcelain disc insulators (the classic stacked disc design)
    \item Damage patches should be CLEARLY VISIBLE (roughly 10-30\% of disc surface, 1-3 patches)
    \item Patches should be matte/faded/discolored with visible contrast to the base glaze
    \item MANDATORY: Each damage patch MUST have a visible thin white/\allowbreak very light-colored edge
    \item Keep the object, lighting, and background consistent with the reference
\end{itemize}

DO NOT:
\begin{itemize}
    \item Generate polymer/silicone/rubber insulators
    \item Show peeling, flaking, raised deposits, or crusty/3D texture
    \item Show chips, cracks, or missing pieces (that's shell damage, not glaze)
\end{itemize}
\end{quote}

\textbf{Shell Damage (Single-Reference):}

\begin{quote}
\small
Based on the provided reference image of a CERAMIC PORCELAIN disc insulator, generate a realistic variation for a dataset.

SHELL DAMAGE means a visible chunk of the rim is missing (broken/chipped porcelain). The damage must be OBVIOUS at first glance. Match the SIZE seen in the reference: roughly 30-70\% of the rim area on affected discs.

REQUIREMENTS:
\begin{itemize}
    \item MUST be ceramic/porcelain disc insulators
    \item Damage must remove a clearly visible chunk of the rim (about 30-70\% on affected discs)
    \item Allow 1-5 affected discs, but avoid catastrophic destruction
    \item Fracture surfaces should look like clean, smooth porcelain (white or very light-colored)
    \item Keep wound edges continuous and ceramic-like; avoid debris or dirt on the fracture
    \item Keep the object, lighting, and background consistent with the reference
\end{itemize}

DO NOT:
\begin{itemize}
    \item Make the damage tiny or hairline; it must be clearly visible
    \item Obliterate entire discs; keep the string intact
    \item Generate polymer/silicone insulators
\end{itemize}
\end{quote}

\subsection{Dual-Reference Prompts}

\textbf{Glaze Damage (V1 - Basic):}

\begin{quote}
\small
I'm providing TWO reference images showing CERAMIC PORCELAIN disc insulators with glaze damage. These are classic PORCELAIN disc-shaped insulators used on power lines---NOT polymer/silicone insulators.

GLAZE DAMAGE is a COLOR CHANGE in the ceramic surface itself---areas where the shiny glaze has become matte, faded, or discolored. The damaged areas are FLUSH with the surface (not raised, not peeling).

Study the reference images carefully, then generate a NEW photo of a ceramic disc insulator with similar glaze damage.

REQUIREMENTS:
\begin{itemize}
    \item MUST be ceramic/porcelain disc insulators (the classic stacked disc design)
    \item Damage patches should be CLEARLY VISIBLE (10-25\% of disc surface)
    \item VARY the insulator color: use brown, reddish-brown, gray, white, OR blue-gray
    \item Use your own background, angle, and lighting
\end{itemize}

DO NOT:
\begin{itemize}
    \item Generate polymer/silicone/rubber insulators
    \item Show peeling, flaking, or raised deposits
    \item Show chips, cracks, or missing pieces (that's shell damage, not glaze)
    \item Copy the reference images---create something new
    \item Always use brown/red colors---vary the insulator color
\end{itemize}
\end{quote}

\textbf{Glaze Damage (V2 - With White Edge):}

\begin{quote}
\small
I'm providing TWO reference images showing CERAMIC PORCELAIN disc insulators with glaze damage. These are classic PORCELAIN disc-shaped insulators used on power lines---NOT polymer/silicone insulators.

GLAZE DAMAGE is a COLOR CHANGE in the ceramic surface itself---areas where the shiny glaze has become matte, faded, or discolored. The damaged areas are FLUSH with the surface (not raised, not peeling).

CRITICAL: Real glaze damage shows a DISTINCT LIGHT/WHITE EDGE or border around the damaged patch. This looks like a thin white or very light-colored intermediate layer between the outer colored glaze (red/brown) and the inner white ceramic. Study the reference images---you MUST see this white edge pattern around the damage patches.

Study the reference images carefully, then generate a NEW photo of a ceramic disc insulator with similar glaze damage.

REQUIREMENTS:
\begin{itemize}
    \item MUST be ceramic/porcelain disc insulators (the classic stacked disc design)
    \item Damage patches should be CLEARLY VISIBLE (roughly 10-30\% of disc surface, 1-3 patches)
    \item Patches should be matte/faded/discolored with visible contrast to the base glaze; edges can be soft/irregular but must stay FLUSH
    \item MANDATORY: Each damage patch MUST have a visible thin white/\allowbreak very light-colored edge or border around it, showing the intermediate layer between glaze and ceramic
    \item VARY the insulator color: use brown, reddish-brown, gray, white, OR blue-gray
    \item Use your own background, angle, and lighting
\end{itemize}

DO NOT:
\begin{itemize}
    \item Generate polymer/silicone/rubber insulators
    \item Show peeling, flaking, raised deposits, or crusty/3D texture
    \item Show chips, cracks, or missing pieces (that's shell damage, not glaze)
    \item Copy the reference images---create something new
\end{itemize}
\end{quote}

\textbf{Shell Damage (V1 - Basic):}

\begin{quote}
\small
I'm providing TWO reference images showing CERAMIC PORCELAIN disc insulators with shell damage. These are classic PORCELAIN disc-shaped insulators used on power lines.

SHELL DAMAGE means chips or cracks at the rim/edge of the porcelain disc where a piece has broken off. Look at the SIZE of the damage in the reference images and MATCH that size---the damage should be CLEARLY VISIBLE.

Study the reference images carefully for the SIZE and SCALE of the damage, then generate a NEW photo of a ceramic disc insulator with similar shell damage.

REQUIREMENTS:
\begin{itemize}
    \item MUST be ceramic/porcelain disc insulators
    \item Damage size should MATCH what's shown in references---clearly visible chips
    \item Typically affects 1-2 discs in a string
    \item VARY the insulator color: use brown, reddish-brown, gray, white, OR blue-gray
    \item Use your own background, angle, and lighting
\end{itemize}

DO NOT:
\begin{itemize}
    \item Make the damage TOO SMALL---match the reference damage size
    \item Show the cement interior heavily exposed
    \item Generate polymer/silicone insulators
    \item Copy the reference images---create something new
    \item Always use brown/red colors---vary the insulator color
\end{itemize}
\end{quote}

\textbf{Shell Damage (V2 - Enhanced Detail):}

\begin{quote}
\small
I'm providing TWO reference images showing CERAMIC PORCELAIN disc insulators with shell damage. These are classic PORCELAIN disc-shaped insulators used on power lines.

SHELL DAMAGE means a visible chunk of the rim is missing (broken/chipped porcelain). The damage must be OBVIOUS at first glance. Match the SIZE seen in the references: roughly 30-70\% of the rim area on affected discs, and in some cases up to about 80\% for a few discs where most of the rim is gone and the core shows.

Study the reference images carefully for SIZE and SCALE, then generate a NEW photo of a ceramic disc insulator with similar shell damage.

REQUIREMENTS:
\begin{itemize}
    \item MUST be ceramic/porcelain disc insulators
    \item Damage must remove a clearly visible chunk of the rim (about 30-70\% on affected discs; occasionally up to $\sim$80\%)
    \item Allow 1-5 affected discs (occasionally up to 7), but avoid catastrophic destruction
    \item Fracture surfaces should look like clean, smooth porcelain (white or very light-colored, uniform), not rough or porous
    \item Keep the wound edges continuous and ceramic-like; avoid debris or dirt on the fracture
    \item VARY the insulator color: use brown, reddish-brown, gray, white, OR blue-gray
    \item Use your own background, angle, and lighting
\end{itemize}

DO NOT:
\begin{itemize}
    \item Make the damage tiny or hairline; it must be clearly visible when viewing the whole string
    \item Obliterate entire discs or remove most of multiple discs; a few discs can have large rim loss with core visible, but keep the string intact
    \item Generate polymer/silicone insulators
    \item Copy the reference images---create something new
\end{itemize}
\end{quote}

\bibliographystyle{elsarticle-num}
\bibliography{references}

\end{document}